\documentclass{article}
\PassOptionsToPackage{numbers}{natbib}

\usepackage[final]{neurips_2024}
\usepackage[utf8]{inputenc}
\usepackage[T1]{fontenc}
\usepackage{hyperref}
\usepackage{url}
\usepackage{booktabs}
\usepackage{amsfonts}
\usepackage{nicefrac}
\usepackage{microtype}
\usepackage{xcolor}
\usepackage{graphicx}
\usepackage{comment}
\usepackage{multirow}
\usepackage{amsmath}
\usepackage{algorithm}
\usepackage{algpseudocode}
\usepackage{wrapfig}
\usepackage{placeins}
[square,numbers,sort&compress]

\title{On Adversarial Robustness of Language Models in Transfer Learning}

\author{
Bohdan Turbal$^{1}$ \quad Anastasiia Mazur$^{2}$ \quad Jiaxu Zhao$^{3}$ \quad
Mykola Pechenizkiy$^{3}$ \\
$^1$Taras Shevchenko National University of Kyiv \quad \\ $^2$National Technical University of Ukraine
“Igor Sikorsky Kyiv Polytechnic Institute”  \\ \quad $^3$Eindhoven University of Technology\\
\texttt{bogdan.turbal.y@gmail.com, anastasiyamazur.wm@gmail.com} \\ 
\texttt{\{j.zhao, m.pechenizkiy\}@tue.nl}
}
\begin{document}

\maketitle

\begin{abstract}
We investigate the adversarial robustness of LLMs in transfer learning scenarios. Through comprehensive experiments on multiple datasets (MBIB Hate Speech, MBIB Political Bias, MBIB Gender Bias) and various model architectures (BERT, RoBERTa, GPT-2, Gemma, Phi), we reveal that transfer learning, while improving standard performance metrics, often leads to increased vulnerability to adversarial attacks. Our findings demonstrate that larger models exhibit greater resilience to this phenomenon, suggesting a complex interplay between model size, architecture, and adaptation methods. Our work highlights the crucial need for considering adversarial robustness in transfer learning scenarios and provides insights into maintaining model security without compromising performance. These findings have significant implications for the development and deployment of LLMs in real-world applications where both performance and robustness are paramount.
\end{abstract}

\section{Introduction}

Large Language Models (LLMs) have become pivotal in natural language processing (NLP), demonstrating remarkable performance across various tasks. Transfer learning, a technique leveraging pre-trained models for new tasks, has significantly contributed to this success \cite{gururangan2020don}. However, the intersection of transfer learning and adversarial robustness in LLMs remains understudied, presenting a critical gap in understanding models' security and reliability.

While transfer learning efficiently applies pre-trained models to new domains, it may inadvertently introduce or amplify vulnerabilities to adversarial attacks. These attacks pose significant threats to model deployment in real-world scenarios. Despite the widespread adoption of transfer learning, there is a notable lack of comprehensive research on how these adapted models perform against adversarial attacks. 

Previous studies have primarily focused on the robustness of models in their initial training or fine-tuning stages \cite{schwinn2023adversarial} \cite{yang2024assessing} \cite{bueno2024}, often in controlled environments. This approach overlooks the potential risks emerging from more complex training sequences, particularly those involving multiple pre-training stages as in transfer learning scenarios. The impact of transfer learning on model robustness is nuanced and multifaceted. While some research suggests that post-fine-tuning can lead to decreased robustness \cite{yang2024assessing}, other findings indicate that incorporating additional data from the target dataset can enhance robustness \cite{xiong2024all}. However, in transfer learning scenarios involving pretraining on related but distinct domains, the impact on robustness becomes more complex and warrants careful investigation.

Our study aims to address the following key research questions:

\textbf{RQ1}: How does transfer learning affect LLMs' performance and robustness overall?

\textbf{RQ2}: How do specific model sizes, architectures, and training procedures influence the transfer learning effect on robustness?

\textbf{RQ3}: What are the potential challenges in real-world scenarios based on these findings?

\textbf{RQ4}: Can adversarial training during transfer learning improve robustness for long sequences, and how does initial dataset specialization influence the robustness-accuracy trade-off?
% 1. How does transfer learning affect the adversarial robustness of LLMs across different model architectures and sizes?

% 2. What strategies can be employed to maintain or improve adversarial robustness during transfer learning while preserving performance?

% 3. How do model characteristics (e.g., size, architecture) influence the trade-off between robustness and accuracy in transfer learning scenarios?
% Our findings reveal a complex relationship between transfer learning, model performance, and adversarial robustness. We observe that while transfer learning often improves standard performance metrics, it can lead to increased vulnerability to adversarial attacks, particularly in smaller models. Larger models, however, demonstrate greater resilience to this phenomenon. These insights have significant implications for the development and deployment of LLMs in security-critical applications.

Our contributions are as follows:
\begin{enumerate}
\item We conducted comprehensive experiments to evaluate the robustness of LLMs against adversarial attacks in transfer learning, revealing that transfer learning often increases vulnerability to adversarial attacks, even when improving standard performance metrics.

\item We provide a detailed analysis of how model characteristics influence robustness in transfer learning scenarios, demonstrating that larger models show significantly greater resilience to increases in vulnerability. This finding is contextualized within a broader examination of how different model architectures (e.g., GPT \citep{radford2019language}, BERT \cite{devlin2018bert}, RoBERTa \cite{liu2019roberta}) and transfer learning techniques (such as LoRA \cite{hu2021lora}) impact robustness, revealing a complex interplay between model size, architecture, and adaptation methods in determining a model's security against adversarial attacks.

\item We provide insights into the trade-off between model robustness and accuracy by conducting experiments with training on perturbed data. This helped us understand the balance between maintaining robustness and preserving model performance.
\end{enumerate}

\section{Experimental Design}
% \subsection{Research Questions}
% Our experiments were designed to address the following research questions:
We focus on the classification task of detecting biased text. We selected the MBIB Hate Speech \cite{wessel2023introducing}, MBIB Political Bias, and MBIB Gender Bias datasets for their relevance to real-world applications and the importance of robustness in these domains. 

% Figure \ref{fig:experiment_setup} illustrates our experimental setup. 

\subsection{Datasets}
We selected three distinct datasets that share a common theme of detecting bias in textual data but address different subdomains within this broader context. This choice allows us to meaningfully explore the impact of transfer learning, as it involves transferring knowledge across related yet distinct types of biases. Each dataset has 2 classes: biased and non-biased, and is balanced. In general, the ability to accurately detect and mitigate various forms of bias is crucial to develop fair and ethical AI systems that can be safely deployed in diverse real-world applications \cite{hendrycks2023overview}. Data sets are selected from \cite{wessel2023introducing} and are as follows: 

\textbf{MBIB Hate Speech (HS)} focuses on identifying hate speech in text. 

\textbf{MBIB Political Bias (PB)} is used to detect political bias in textual data. 

\textbf{MBIB Gender Bias (GB)} helps evaluate the model's ability to recognize gender bias in text.

For each dataset, we create a large training set (12,000) for pre-training, a small training set (600) for target task fine-tuning, a validation set (1,000) and a test set (1,000).

\subsection{Evaluation Metrics}
We use the following metrics to assess model performance and robustness:

\textbf{Original Accuracy (OAcc):} Main usual metric to evaluate the performance of the model in classification.

\textbf{Attack Success Rate (ASR):} Percentage of True Positive and True Negative examples that were hacked by the attack, this metric can serve as a basic evaluation of the robustness of the model.

\textbf{Accuracy Under Attack (AUA): } The accuracy of the model after attack. This metric can be considered a `safety' metric for the model. For instance, if the model's accuracy (Acc) significantly increases while the Attack Success Rate (ASR) only mildly increases, the AUA may show improvement even though the model has become less robust overall.

\subsection{Parameters Setting}

For the pre-training phase, we trained the models for 1 epoch on the larger subset of the dataset. During the fine-tuning phase on the target dataset, the models were trained for up to 6 epochs, with the best model selected based on validation set accuracy and, in cases where training loss differed drastically at the baseline epoch count, trained to achieve roughly comparable training loss on the target dataset across model sequences to ensure fair comparison. We used the Adam optimizer, adjusting the learning rate between $5 \times 10^{-6}$ and $4 \times 10^{-4}$ depending on the specific model, to ensure optimal convergence during training.

\subsection{Training Procedure}
We measured the performance under two conditions:

\textbf{Usual Fine-Tuning:} Fine-tuning the model directly on the target dataset.

\textbf{Transfer Learning:} Fine-tuning the model on one large dataset followed by transfer learning to the target task using the smaller dataset.

\begin{figure}[ht]
\centering
\includegraphics[width=0.7\textwidth]{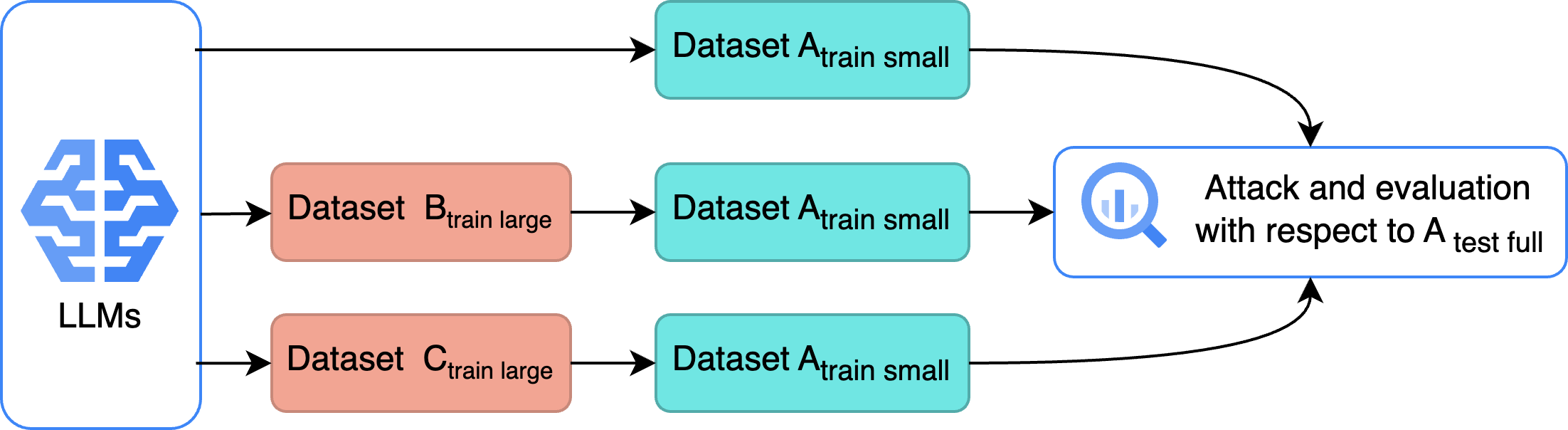}
\caption{Experiment setup, where we compare LLM's properties in additional Transfer learning setup and just target dataset fine-tuning.}
\label{fig:experiment_setup}
\end{figure}

This setup allowed us to measure the influence of each training sequence on both accuracy and adversarial robustness, providing insights into the trade-offs involved in using transfer learning for LLMs in classification tasks aimed at detecting biased versus non-biased text. The overall experiment setup is displayed in Figure \ref{fig:experiment_setup}.
%Other additional experimental setups are shown in %Appendix \ref{app: experiment setup}.

\subsection{Experimental Setup}\label{sec:experimental_setup}
In our experiments, we conducted two main types of experiments:

\textbf{Standard Transfer Learning}: Fine-tuning models on one dataset followed by transfer to another.

\textbf{Adversarial Training with Transfer Learning}: Incorporating adversarial examples (10\% of training data) during the transfer learning process.
\subsection{Models}

We evaluated a range of LLMs to assess the impact of model size and architecture on robustness: BERT Base (110M), BERT Large (340M), RoBERTa Base (125M), RoBERTa Large (355M), GPT-2 (117M), GPT-2 Medium (345M), and GPT-2 Large (762M) and also large models like Gemma 2b (2B), Phi-2 (2.7B), and GPT2-XL (1.5B).

% We evaluated a range of LLMs to assess the impact of model size and architecture on robustness: BERT Base (110M parameters), BERT Large (340M parameters), RoBERTa Base (125M parameters), RoBERTa Large (355M parameters), GPT-2 (117M parameters), GPT-2 Medium (345M parameters), and GPT-2 Large (762M parameters) and also large models like Gemma 2b, Phi-2, and GPT2-XL.

\subsection{Adversarial Attack Methods}
We employ two attack methods in our experiments:

\textbf{TextFooler} \cite{jin2020bert}: A word-level adversarial attack method for text classification. It uses word deletion impact for importance ranking, word embeddings for synonyms, and Universal Sentence Encoder for semantic similarity constraints.

\textbf{A2T} \cite{yoo2021towards}: A computationally efficient adversarial attack method. It uses gradient-based word importance ranking, counter-fitted word embeddings for synonyms, and DistilBERT for semantic similarity constraints.

\section{Results and Analysis}
Our experiments yielded several key insights into the impact of transfer learning on the robustness and performance of LLMs. We'll discuss our findings in relation to each research question.
\begin{table}[ht]
\centering
\scriptsize
\caption{Performance and safety metrics comparison for models with and without transfer learning, averaged across attack methods (Text Fooler and A2T) and individual sequences ending with the target dataset. OAcc - Original Accuracy, ASR - Attack Sucess Rate, AUA - Accuracy Under Attack, $\Delta$ (\%) - relative change. }

\begin{tabular}{c|c|ccc|ccc|cc}
\hline
\multirow{2}{*}{\textbf{Model}} & \multirow{2}{*}{\textbf{Target dataset}} & \multicolumn{3}{c|}{Without Transfer Learning} & \multicolumn{3}{c|}{Transfer Learning} & \multicolumn{2}{c}{Results} \\ \cline{3-10}
 &  & OAcc& ASR & AUA & OAcc& ASR & AUA & $\Delta$ OAcc(\%) & $\Delta$ ASR (\%) \\ \hline
\multirow{3}{*}{GPT-2} & Gender bias & 0.763 & 0.397 & 0.460 & 0.791 & \textbf{0.478} & 0.413 & 3.67 & 20.40 \\ 
 & Political bias & 0.650 & 0.457 & 0.357 & 0.659 & \textbf{0.514} & 0.321 & 1.38 & 12.47 \\ 
 & Hate speech & 0.798 & 0.463 & 0.428 & 0.796 & \textbf{0.472} & 0.420 & -0.25 & 1.94 \\ \hline
\multirow{3}{*}{GPT-2-medium} & Gender bias & 0.743 & 0.452 & 0.408 & 0.806 & \textbf{0.491} & 0.410 & 8.48 & 8.63 \\
 & Political bias & 0.657 & 0.350 & 0.427 & 0.677 & \textbf{0.445} & 0.375 & 3.04 & 27.14 \\ 
 & Hate speech & 0.823 & 0.413 & 0.483 & 0.824 & \textbf{0.423} & 0.475 & 0.12 & 2.42 \\ \hline
\multirow{3}{*}{GPT-2-large} & Gender bias & 0.780 & \textbf{0.460} & 0.421 & 0.789 & 0.456 & 0.429 & 1.15 & -0.87 \\ 
 & Political bias & 0.675 & 0.430 & 0.385 & 0.679 & \textbf{0.444} & 0.378 & 0.59 & 3.26 \\ 
 & Hate speech & 0.832 & 0.437 & 0.468 & 0.823 & \textbf{0.444} & 0.458 & -1.08 & 1.60 \\ \hline
\multirow{3}{*}{BERT} & Gender bias & 0.768 & 0.436 & 0.433 & 0.778 & \textbf{0.558} & 0.344 & 1.30 & 28.00 \\ 
 & Political bias & 0.690 & 0.470 & 0.366 & 0.670 & \textbf{0.516} & 0.324 & -2.90 & 9.79 \\
 & Hate speech & 0.775 & 0.490 & 0.395 & 0.787 & \textbf{0.515} & 0.382 & 1.55 & 5.10 \\ \hline
\multirow{3}{*}{BERT-large} & Gender bias & 0.743 & 0.485 & 0.383 & 0.787 & \textbf{0.513} & 0.383 & 5.9 & 5.7 \\ 
 & Political bias & 0.682 & \textbf{0.500} & 0.341 & 0.678 & 0.495 & 0.342 & -0.59 & -1.00 \\
 & Hate speech & 0.732 & 0.491 & 0.378 & 0.757 & \textbf{0.516} & 0.366 & 3.42 & 5.09 \\ \hline
\multirow{3}{*}{RoBERTa} & Gender bias & 0.807 & 0.465 & 0.432 & 0.791 & \textbf{0.557} & 0.350 & -1.98 & 19.78 \\ 
 & Political bias & 0.680 & 0.348 & 0.443 & 0.677 & \textbf{0.445} & 0.376 & -0.44 & 27.87 \\ 
 & Hate speech & 0.795 & 0.431 & 0.452 & 0.817 & \textbf{0.469} & 0.433 & 2.77 & 8.82 \\ \hline
\multirow{3}{*}{RoBERTa-large} & Gender bias & 0.807 & \textbf{0.503} & 0.401 & 0.817 & 0.463 & 0.438 & 1.2 & -7.95 \\ 
 & Political bias & 0.708 & 0.355 & 0.457 & 0.717 & \textbf{0.474} & 0.377 & 1.27 & 33.5 \\ 
 & Hate speech & 0.825 & 0.360 & 0.528 & 0.827 & \textbf{0.403} & 0.494 & 0.24 & 11.9 \\ \hline

 \multirow{3}{*}{Phi-2 (LoRA)} & Gender bias & 0.803 & 0.514 & 0.390 & 0.815 & \textbf{0.519} & 0.394 & 1.49 & 0.97 \\ 
 & Political bias & 0.745 & \textbf{0.591} & 0.308 & 0.721 & 0.513 & 0.350 & -3.22 & -13.2 \\ 
 & Hate speech & 0.833 & 0.396 & 0.503 & 0.845 & \textbf{0.432} & 0.479 & 1.44 & 9.09 \\ \hline
\multirow{3}{*}{Gemma 2B (LoRA)} & Gender bias & 0.788 & 0.575 & 0.335 & 0.796 & \textbf{0.611} & 0.310 & 1.02 & 6.26 \\
 & Political bias & 0.703 & \textbf{0.677} & 0.226 & 0.685 & 0.587 & 0.279 & -2.56 & -13.29 \\ 
 & Hate speech & 0.765 & 0.426 & 0.439 & 0.774 & \textbf{0.499} & 0.388 & 1.17 & 17.13 \\ \hline
\multirow{3}{*}{GPT-2-xl (LoRA)} & Gender bias & 0.825 & 0.509 & 0.406 & 0.820 & \textbf{0.551} & 0.369 & -0.61 & 8.25 \\ 
 & Political bias & 0.683 & 0.527 & 0.323 & 0.689 & \textbf{0.547} & 0.313 & 0.88 & 3.80 \\ 
 & Hate speech & 0.833 & 0.448 & 0.461 & 0.831 & \textbf{0.465} & 0.444 & -0.24 & 3.79 \\ \hline
\end{tabular}
\label{table:comparison}
\end{table}

\subsection{RQ1: Transfer learning robustness}
We evaluated various LLMs using TextFooler (black-box) and A2T (white-box) adversarial attacks. The results, presented in Table \ref{table:comparison}, reveal a concerning trend:

Increased Vulnerability: In most cases, especially for smaller models, the Attack Success Rate (ASR) increased after transfer learning, regardless of changes in accuracy (OAcc). It suggests that even when models demonstrated enhanced performance in terms of accuracy, their overall robustness against adversarial attacks often decreased. 

Performance-Robustness Trade-off: Even when models showed improved accuracy, their robustness against adversarial attacks often decreased.
For example, on the Hate Speech dataset, GPT-2 experiences a mean 20.4\% increase in ASR alongside a 3.67\% increase in accuracy. This finding raises significant concerns about LLM security, as improvements in accuracy during training might lead developers to overlook other critical parameters like robustness. 

Table \ref{table:asr_increase_horizontal_gemma} presents the percentage of unaveraged sequences with increased ASR, confirming this robustness decline trend. The complete raw data is available in Appendix \ref{app:transf_raw}.

\begin{table}[ht]
\centering
\scriptsize
\caption{Percentage of unaveraged individual training sequences where ASR increase is observed.}
\resizebox{\textwidth}{!}{%
\begin{tabular}{c|c|c|c|c|c|c|c|c|c|c}
\hline
& GPT-2 & GPT-2 & GPT-2 & BERT & BERT- & RoBERTa & RoBERTa- & Phi-2 & Gemma & GPT-2 \\
& & medium & large & & large & & large & & 2B & xl \\ \hline
\textbf{(\%)} & 83.3 & 91.7 & 50.0 & 91.7 & 75.0 & 100.0 & 66.7 & 41.7 & 58.3 & 75.0 \\
\hline
\end{tabular}%
}
\label{table:asr_increase_horizontal_gemma}
\end{table}

\subsubsection{LoRA and Larger Models}

For large models with billions of parameters, we used LoRA due to its efficiency in adapting these models, as conventional fine-tuning often requires extensive computational resources that may not be readily available in typical settings. When applying LoRA to these larger models, we observed mixed results. Some sequences showed decreased robustness, while others demonstrated increased robustness (e.g., political bias dataset for Phi-2 and Gemma 2b), result are presented in Table \ref{table:comparison} and Table \ref{table:asr_increase_horizontal_gemma}.

% When applying LoRA to larger models, we observed mixed results. We used LoRA due to its efficiency in adapting large models with billions of parameters, as conventional fine-tuning often requires extensive computational resources that may not be readily available in typical settings.  Some sequences showed decreased robustness, while others demonstrated increased robustness (e.g., political bias dataset for Phi-2 and Gemma 2b).

% The impact of LoRA on the experimental results is not straightforward. This is because LoRA operates differently from standard fine-tuning in three key ways: 1) it introduces additional parameters rather than fine-tuning existing parts of the original model, 2) it involves random initialization of these parameters, and 3) they constitute a small percentage of all parameters. One potential explanation is that the random initialization of LoRA adapter parameters, as opposed to fine-tuning pre-trained parameters, may lead to different robustness outcomes compared to standard fine-tuning. While transfer learning can still reduce robustness through issues like false memories \cite{gekhman2024does}, shortcut learning \cite{du2023shortcut}, or increased sensitivity to perturbations acquired\cite{wang2023large}, LoRA might not have problems like catastrophic forgetting \cite{luo2023empirical}. This is because LoRA's adapter starts with no pre-existing information that could be distorted or lost during transfer learning.

The impact of LoRA on robustness is complex due to its unique approach: introducing and randomly initializing a small set of additional parameters rather than fine-tuning existing ones. This may lead to different robustness outcomes compared to standard fine-tuning. While transfer learning here can still reduce robustness through issues like false memories \cite{gekhman2024does} or shortcut learning \cite{du2023shortcut}, catastrophic forgetting \cite{luo2023empirical} may not contribute significantly to the results in this specific setting. This is because, with the random initialization of LoRA adapter parameters and the freezing of other parameters, there is no pre-existing information in the adapters that could be distorted or lost during the transfer learning process, thus potentially altering the dynamics of how robustness changes during transfer learning.

\subsection{RQ2: Impact of Model Size, Architecture, and Training Procedures}
We examined how different model sizes, architectures, and training procedures (including LoRA for larger models) influenced the transfer learning effect on robustness.

\subsubsection{Model Size and Architecture Influence}

%[to do] Figures to appendix?

\begin{figure}[ht]
\centering
\begin{minipage}{0.42\textwidth}
    \centering
    \includegraphics[width=\textwidth]{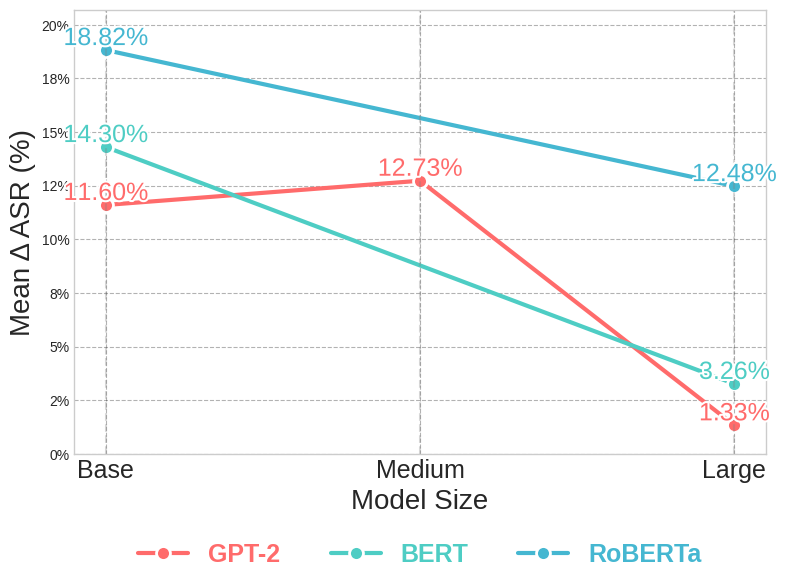}
    \caption{$\Delta$ ASR by model size and architecture.}
    \label{fig:model_size_asr_increase}
\end{minipage}\hfill
\begin{minipage}{0.42\textwidth}
     \centering
     \includegraphics[width=\textwidth]{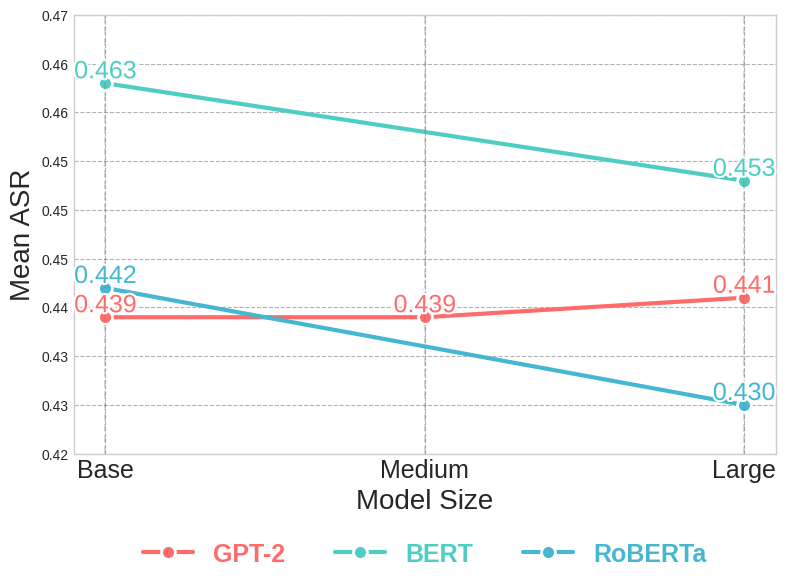}
     \caption{ASR by model size and architecture.}
     \label{fig:asr_by_model}
\end{minipage}
\end{figure}

% \begin{wrapfigure}{r}{0.5\textwidth}
% \centering
% \includegraphics[width=0.47\textwidth]{img/model_size_comp_2.png}
% \caption{$\Delta$ ASR by model size.}
% \label{fig:model_size_asr_increase}
% \end{wrapfigure}
% Larger Models Show Better Resilience: As illustrated in Figure \ref{fig:model_size_asr_increase}, larger models within each family (GPT-2, BERT, RoBERTa) exhibited smaller increases in ASR due to transfer learning.
% \begin{wrapfigure}{r}{0.5\textwidth}
% \centering
% \includegraphics[width=0.48\textwidth]{img/model_size_comp_2.png}
% \caption{$\Delta$ ASR by model size.}
% \label{fig:model_size_asr_increase}
% \end{wrapfigure}

Larger Models Show Better Resilience: As illustrated in Figure \ref{fig:model_size_asr_increase} ,  larger models within each family (GPT-2, BERT, RoBERTa) exhibited smaller increases in ASR due to transfer learning.

Initial ASR Variations: The mean initial ASR (before transfer learning) didn't follow a consistent pattern across model families Figure \ref{fig:asr_by_model}:
\begin{enumerate}
\item Decreased with size in BERT and RoBERTa, but increased with size in GPT-2 models.
% \item Overall ASR range remained similar (0.42 to 0.47) across autoregressive and encoder-based models, indicating that both autoregressive and encoder-based models exhibit comparable levels of robustness against adversarial attacks.

\item Overall ASR range remained similar (0.42 to 0.47) across autoregressive and encoder-based models, indicating that both of them exhibit comparable levels of robustness against adversarial attacks.

\end{enumerate}

\subsection{RQ3: Real-world implications}

As we showed, often ASR increases in parallel to OAcc, which indicates a potential trade-off of using Transfer learning between performance and safety. Often standard metrics like OAcc are prioritized, while other safety metrics are overlooked, leading to vulnerable models being deployed. Based on our findings, we highlight the necessity of applying additional techniques and adversarial testing to mitigate this issue, particularly when fine-tuning smaller models. For Larger Models with LoRA, use transfer learning cautiously, as effects on robustness can vary. 

\subsection{RQ4: Trade-off Between Robustness and Accuracy} \label{trade_off}
\paragraph{Experimental Setup} This experiment explores the balance between robustness and accuracy in LLMs under adversarial attacks during transfer learning. Two methods are compared:

\begin{itemize} 
\item \textbf{Iterative Transfer Learning with Adversarial Attacks}: The model is sequentially trained and evaluated on three datasets (Hate speech (HS), Political bias (PB) and Gender bias (GB)), with a final evaluation across all datasets to track performance over time.
\item \textbf{Adversarial Training with Transfer Learning}: Adversarial samples (10\%) are included during training to enhance robustness, with performance assessed across all datasets.
\end{itemize}

Results in Table\ref{table:adversarial_impact} show how attacks (A2T, TextFooler) impact Original Accuracy (OAcc), Accuracy Under Attack (AUA), and Attack Success Rate (ASR). ``FE'' denotes Final Evaluation, and ``PC'' represents Percent Change relative to earlier evaluations.

\begin{table}[ht]
\caption{Impact of A2T attack on Bert performance, where for convenience of notation: HS, PB and GB are Hate speech, Political bias and Gender bias datasets.}
\scriptsize
\centering
\begin{tabular}{c|ccc|ccc|ccc}
\hline
\textbf{Metrics} & {HS} & {PB} & {GB} & {FE HS} & {FE PB} & {FE GB} & \textbf{$\Delta$ PC HS(\%)} & \textbf{$\Delta$ PC PB(\%)} & \textbf{$\Delta$ PC GB(\%)} \\ \hline
\multicolumn{10}{c}{\textbf{Original Data}} \\ \hline
OAcc   & 78.07 & 69.48 & 70.92 & 69.94 & 62.42 & 70.6  & \textbf{-10.41} & -10.16 & -0.45 \\ 
AUA  & 54.26 & 45.59 & 50.28 & 55.04 & 45.24 & 47.26 & \textbf{1.44} & -0.77 & -6.01 \\ 
ASR  & 30.61 & 34.31 & 30.5  & 21.55 & 29.73 & 33.1  & \textbf{-29.60} & -13.35 & 8.52  \\ \hline
\multicolumn{10}{c}{\textbf{Including Adversarial Training}} \\ \hline
OAcc   & 75.8  & 69.23 & 68.75 & 69.78 & 68.18 & 68.59 & \textbf{-7.94} & -1.52 & -0.23 \\ 
AUA  & 50.25 & 48.24 & 48.38 & 52.51 & 48.93 & 50.26 & 4.50 & \textbf{1.43} & 3.89  \\ 
ASR  & 33.92 & 30.41 & 29.63 & 25.3  & 28.18 & 26.43 & \textbf{-25.41} & -7.33 & -10.80 \\ \hline
\end{tabular}
\label{table:adversarial_impact}
\end{table}

The experiments reveal a trade-off between robustness and accuracy (full results are shown in Table \ref{table:adversarial_combined_impact_textfooller} and Table \ref{table:adversarial_combined_impact_a2t} of Appendix \ref{app: trade}). Adversarial fine-tuning reduces OAcc but significantly boosts AUA and lowers the ASR, especially on the first dataset. Early exposure to adversarial examples enables the model to build strong defense mechanisms, improving its resistance to attacks despite a decline in OAcc.

Introducing adversarial samples during training enhances overall performance, leading to a more robust model. While accuracy on the first dataset decreases, the model's ability to withstand attacks improves, indicating a balanced adaptation between accuracy and robustness over time.

\section{Conclusion}
Our research contributes to the understanding of the adversarial robustness of LLMs in the context of transfer learning. Our empirical analysis reveals nuanced dynamics in the relationship between traditional performance metrics, such as accuracy, and the robustness of models against adversarial attacks. Interestingly, we observed instances where improvements in conventional metrics were accompanied by a decrease in adversarial robustness, suggesting a potential trade-off between performance enhancement and vulnerability to adversarial manipulations. This counterintuitive finding underscores the complexity of model behavior in transfer learning scenarios and raises questions about the underlying causes, which may include phenomena such as catastrophic forgetting or the acquisition of misleading ``false memories'' during pre-training. Notably, our results indicate that larger models may exhibit a reduced susceptibility to this trend, hinting at an inherent robustness associated with scale.

\section*{Acknowledgments}
This research was supported in part by a grant from the Simons Foundation (SFARI award \#1280457, JS) and by the NYU Center for Responsible AI through the RAI for Ukraine program. We thank the support of EU European Defence Fund Project KOIOS (EDF-2021-DIGIT-R-FL-KOIOS).

\vspace{50pt}

\bibliography{iclr2024_conference}
\bibliographystyle{iclr2024_conference}

% \bibliography{iclr2024_conference}
% \bibliographystyle{iclr2024_conference}

\appendix

\section{Social Impact Statement}\label{app: social statement}

Our research rigorously examines the balance between performance enhancements and security vulnerabilities in large language models (LLMs) using transfer learning. This analysis has highlighted the need for training methodologies that prioritize both model effectiveness and security.

As LLMs become more common in sectors like healthcare, finance, and public services, it is crucial to protect these systems from sophisticated adversarial threats. Our findings show that while transfer learning can improve model performance, it can also introduce and magnify vulnerabilities that malicious actors could exploit, necessitating a reevaluation of current training practices.

We advocate for incorporating comprehensive adversarial training and robustness assessments during the AI development. By adopting these practices, developers can better manage the trade-offs between accuracy and security, ensuring that improvements in LLM capabilities do not compromise their defense.

Our study reveals interesting nuances in the interaction between transfer learning, performance, and security. We observed instances where transfer learning not only contributed to performance improvements but also bolstered the models' defenses against adversarial attacks under certain conditions. These insights suggest that transfer learning, when applied thoughtfully, might offer opportunities to simultaneously enhance both the effectiveness and the security of LLMs, meriting deeper investigation into these phenomena.

\section{Transfer learning experiments raw data} \label{app:transf_raw}
While Table \ref{table:comparison} presented averaged results, the raw data provides more granular insights into the behavior of different model architectures and attack methods.

Table \ref{table:gpt2_attack_comparison} through Table \ref{table:gpt2_xl_attack_comparison} demonstrate an inverse relationship between model size and vulnerability to adversarial attacks post-transfer learning within the GPT-2 family. GPT-2 exhibits ASR increases up to 80\% for some sequences, whereas GPT-2 XL's maximum ASR increase is approximately 22\%.

Table \ref{table:bert_attack_comparison} and Table \ref{table:bert_large_attack_comparison} show that BERT models, despite their bidirectional architecture, display vulnerability patterns similar to GPT-2, with BERT Large showing marginally improved robustness.

RoBERTa models (Table \ref{table:roberta_attack_comparison} and Table \ref{table:roberta_large_attack_comparison}) exhibit an noteworthy characteristic: while generally more robust than BERT, they still incur significant ASR increases, particularly against the a2t attack. This suggests that RoBERTa's enhanced pretraining does not necessarily confer improved adversarial robustness in transfer learning scenarios.

The results for Phi-2 and Gemma 2B (Table \ref{table:phi2_attack_comparison} and Table \ref{table:gemma_2b_attack_comparison}) are particularly noteworthy. These LoRA-tuned models show highly variable results, with some sequences demonstrating improved robustness post-transfer. This variability indicates a complex interaction between LoRA's adaptation mechanism and adversarial vulnerability, warranting further investigation.

These raw results not only corroborate our main findings but also elucidate the nuanced impact of model architecture, size, and fine-tuning method on adversarial robustness in transfer learning contexts.

\begin{table}[ht]
\centering
\caption{Performance metrics for a2t attack and text fooler attack on GPT-2.}
\small
\begin{tabular}{r|ccc|ccc}
\hline
\textbf{(Dataset sequence)} & \textbf{OAcc} & \textbf{ASR} & \textbf{AUA} & \textbf{$\Delta$ OAcc} & \textbf{$\Delta$ ASR} & \textbf{$\Delta$ AUA} \\ \hline
\multicolumn{7}{c}{\textbf{a2t attack}} \\ \hline
None -> GB & 0.763 & 0.164 & 0.638 & (baseline) & (baseline) & (baseline) \\
PB -> GB & 0.798 & 0.298 & 0.560 & 4.59\% & 81.66\% & -12.16\% \\
HS -> GB & 0.785 & 0.236 & 0.600 & 2.95\% & 43.76\% & -5.88\% \\
GB -> PB & 0.650 & 0.396 & 0.393 & 0.00\% & 13.19\% & -7.10\% \\
None -> PB & 0.650 & 0.350 & 0.423 & (baseline) & (baseline) & (baseline) \\
HS -> PB & 0.668 & 0.427 & 0.383 & 2.69\% & 21.99\% & -9.47\% \\
GB -> HS & 0.808 & 0.424 & 0.465 & 1.25\% & 10.00\% & -5.10\% \\
PB -> HS & 0.785 & 0.411 & 0.463 & -1.57\% & 6.55\% & -5.61\% \\
None -> HS & 0.798 & 0.386 & 0.490 & (baseline) & (baseline) & (baseline) \\ \hline
\multicolumn{4}{r|}{Average} & 1.653301946 & 29.52457272 & -7.553591456 \\ \hline
\multicolumn{7}{c}{\textbf{text fooler attack}} \\ \hline
None -> GB & 0.763 & 0.630 & 0.283 & (baseline) & (baseline) & (baseline) \\
PB -> GB & 0.798 & 0.708 & 0.232 & 4.59\% & 12.30\% & -18.24\% \\
HS -> GB & 0.785 & 0.669 & 0.260 & 2.95\% & 6.20\% & -8.24\% \\
GB -> PB & 0.650 & 0.591 & 0.272 & 0.00\% & 4.75\% & -6.86\% \\
None -> PB & 0.650 & 0.565 & 0.292 & (baseline) & (baseline) & (baseline) \\
HS -> PB & 0.668 & 0.642 & 0.237 & 2.69\% & 13.75\% & -18.86\% \\
GB -> HS & 0.808 & 0.534 & 0.378 & 1.25\% & -1.09\% & 3.18\% \\
PB -> HS & 0.785 & 0.520 & 0.375 & -1.57\% & -3.61\% & 2.27\% \\
None -> HS & 0.798 & 0.540 & 0.367 & (baseline) & (baseline) & (baseline) \\ \hline
\multicolumn{4}{r|}{Average} & 1.653301946 & 5.382223872 & -7.788388083 \\ \hline
\end{tabular}
\label{table:gpt2_attack_comparison}
\end{table}

\begin{table}[ht]
\centering
\caption{Performance metrics for a2t attack and text fooler attack on GPT-2 medium.}
\small
\begin{tabular}{r|ccc|ccc}
\hline
\textbf{(Dataset sequence)} & \textbf{OAcc} & \textbf{ASR} & \textbf{AUA} & \textbf{$\Delta$ OAcc} & \textbf{$\Delta$ ASR} & \textbf{$\Delta$ AUA} \\ \hline
\multicolumn{7}{c}{\textbf{a2t attack}} \\ \hline
None -> GB & 0.743 & 0.242 & 0.563 & (baseline) & (baseline) & (baseline) \\
PB -> GB & 0.805 & 0.313 & 0.553 & 8.30\% & 29.10\% & -1.78\% \\
HS -> GB & 0.807 & 0.271 & 0.588 & 8.52\% & 11.77\% & 4.44\% \\
GB -> PB & 0.667 & 0.298 & 0.468 & 1.52\% & 26.04\% & -6.64\% \\
None -> PB & 0.657 & 0.236 & 0.502 & (baseline) & (baseline) & (baseline) \\
HS -> PB & 0.687 & 0.323 & 0.465 & 4.57\% & 36.76\% & -7.31\% \\
GB -> HS & 0.833 & 0.358 & 0.535 & 1.21\% & 5.90\% & -1.83\% \\
PB -> HS & 0.815 & 0.335 & 0.542 & -1.01\% & -0.79\% & -0.61\% \\
None -> HS & 0.823 & 0.338 & 0.545 & (baseline) & (baseline) & (baseline) \\ \hline
\multicolumn{4}{r|}{Average} & 3.851657201 & 18.13081029 & -2.28954161 \\ \hline
\multicolumn{7}{c}{\textbf{text fooler attack}} \\ \hline
None -> GB & 0.743 & 0.661 & 0.252 & (baseline) & (baseline) & (baseline) \\
PB -> GB & 0.805 & 0.698 & 0.243 & 8.30\% & 5.49\% & -3.31\% \\
HS -> GB & 0.807 & 0.684 & 0.255 & 8.52\% & 3.39\% & 1.32\% \\
GB -> PB & 0.667 & 0.555 & 0.297 & 1.52\% & 19.49\% & -15.64\% \\
None -> PB & 0.657 & 0.464 & 0.352 & (baseline) & (baseline) & (baseline) \\
HS -> PB & 0.687 & 0.604 & 0.272 & 4.57\% & 30.12\% & -22.75\% \\
GB -> HS & 0.833 & 0.510 & 0.408 & 1.21\% & 4.54\% & -3.16\% \\
PB -> HS & 0.815 & 0.489 & 0.417 & -1.01\% & 0.18\% & -1.19\% \\
None -> HS & 0.823 & 0.488 & 0.422 & (baseline) & (baseline) & (baseline) \\ \hline
\multicolumn{4}{r|}{Average} & 3.851657201 & 10.53609032 & -7.453867774 \\ \hline
\end{tabular}
\label{table:gpt2_medium_attack_comparison}
\end{table}

\begin{table}[ht]
\centering
\caption{Performance metrics for a2t attack and text fooler attack on GPT-2 large.}
\small
\begin{tabular}{r|ccc|ccc}
\hline
\textbf{(Dataset sequence)} & \textbf{OAcc} & \textbf{ASR} & \textbf{AUA} & \textbf{$\Delta$ OAcc} & \textbf{$\Delta$ ASR} & \textbf{$\Delta$ AUA} \\ \hline
\multicolumn{7}{c}{\textbf{a2t attack}} \\ \hline
None -> GB & 0.780 & 0.254 & 0.582 & (baseline) & (baseline) & (baseline) \\
PB -> GB & 0.770 & 0.190 & 0.623 & -1.28\% & -25.09\% & 7.16\% \\
HS -> GB & 0.808 & 0.330 & 0.542 & 3.63\% & 29.74\% & -6.88\% \\
GB -> PB & 0.685 & 0.326 & 0.462 & 1.48\% & 4.80\% & -0.72\% \\
None -> PB & 0.675 & 0.311 & 0.465 & (baseline) & (baseline) & (baseline) \\
HS -> PB & 0.673 & 0.285 & 0.482 & -0.25\% & -8.50\% & 3.58\% \\
GB -> HS & 0.825 & 0.382 & 0.510 & -0.80\% & -0.25\% & -0.65\% \\
PB -> HS & 0.820 & 0.384 & 0.505 & -1.40\% & 0.36\% & -1.62\% \\
None -> HS & 0.832 & 0.383 & 0.513 & (baseline) & (baseline) & (baseline) \\ \hline
\multicolumn{4}{r|}{Average} & 0.230097739 & 0.1760995019 & 0.1468648652 \\ \hline
\multicolumn{7}{c}{\textbf{text fooler attack}} \\ \hline
None -> GB & 0.780 & 0.667 & 0.260 & (baseline) & (baseline) & (baseline) \\
PB -> GB & 0.770 & 0.662 & 0.260 & -1.28\% & -0.65\% & 0.00\% \\
HS -> GB & 0.808 & 0.641 & 0.290 & 3.63\% & -3.81\% & 11.54\% \\
GB -> PB & 0.685 & 0.567 & 0.297 & 1.48\% & 3.42\% & -2.73\% \\
None -> PB & 0.675 & 0.548 & 0.305 & (baseline) & (baseline) & (baseline) \\
HS -> PB & 0.673 & 0.597 & 0.272 & -0.25\% & 8.83\% & -10.93\% \\
GB -> HS & 0.825 & 0.489 & 0.422 & -0.80\% & -0.43\% & -0.39\% \\
PB -> HS & 0.820 & 0.520 & 0.393 & -1.40\% & 5.98\% & -7.09\% \\
None -> HS & 0.832 & 0.491 & 0.423 & (baseline) & (baseline) & (baseline) \\ \hline
\multicolumn{4}{r|}{Average} & 0.230097739 & 2.222775906 & -1.599892317 \\ \hline
\end{tabular}
\label{table:gpt2_large_attack_comparison}
\end{table}

\begin{table}[ht]
\centering
\caption{Performance metrics for a2t attack and text fooler attack on BERT.}
\small
\begin{tabular}{r|ccc|ccc}
\hline
\textbf{(Dataset sequence)} & \textbf{OAcc} & \textbf{ASR} & \textbf{AUA} & \textbf{$\Delta$ OAcc} & \textbf{$\Delta$ ASR} & \textbf{$\Delta$ AUA} \\ \hline
\multicolumn{7}{c}{\textbf{a2t attack}} \\ \hline
None -> GB & 0.768 & 0.254 & 0.573 & (baseline) & (baseline) & (baseline) \\
PB -> GB & 0.765 & 0.390 & 0.467 & -0.43\% & 53.66\% & -18.60\% \\
HS -> GB & 0.792 & 0.354 & 0.512 & 3.04\% & 39.36\% & -10.76\% \\
GB -> PB & 0.697 & 0.390 & 0.425 & 0.97\% & 8.35\% & -3.77\% \\
None -> PB & 0.690 & 0.360 & 0.442 & (baseline) & (baseline) & (baseline) \\
HS -> PB & 0.643 & 0.394 & 0.390 & -6.76\% & 9.41\% & -11.70\% \\
GB -> HS & 0.792 & 0.436 & 0.447 & 2.15\% & 3.39\% & -0.37\% \\
PB -> HS & 0.783 & 0.457 & 0.425 & 1.08\% & 8.53\% & -5.20\% \\
None -> HS & 0.775 & 0.422 & 0.448 & (baseline) & (baseline) & (baseline) \\ \hline
\multicolumn{4}{r|}{Average} & 0.005290313116 & 20.44898997 & -8.401395235 \\ \hline
\multicolumn{7}{c}{\textbf{text fooler attack}} \\ \hline
None -> GB & 0.768 & 0.618 & 0.293 & (baseline) & (baseline) & (baseline) \\
PB -> GB & 0.765 & 0.773 & 0.173 & -0.43\% & 25.10\% & -40.91\% \\
HS -> GB & 0.792 & 0.716 & 0.225 & 3.04\% & 15.78\% & -23.30\% \\
GB -> PB & 0.697 & 0.641 & 0.250 & 0.97\% & 10.60\% & -13.79\% \\
None -> PB & 0.690 & 0.580 & 0.290 & (baseline) & (baseline) & (baseline) \\
HS -> PB & 0.643 & 0.640 & 0.232 & -6.76\% & 10.38\% & -20.11\% \\
GB -> HS & 0.792 & 0.552 & 0.355 & 2.15\% & -1.35\% & 3.90\% \\
PB -> HS & 0.783 & 0.617 & 0.300 & 1.08\% & 10.35\% & -12.20\% \\
None -> HS & 0.775 & 0.559 & 0.342 & (baseline) & (baseline) & (baseline) \\ \hline
\multicolumn{4}{r|}{Average} & 0.005290313116 & 11.81101791 & -17.73421239 \\ \hline
\end{tabular}
\label{table:bert_attack_comparison}
\end{table}

\begin{table}[ht]
\centering
\caption{Performance metrics for a2t attack and text fooler attack on BERT-large.}
\small
\begin{tabular}{r|ccc|ccc}
\hline
\textbf{(Dataset sequence)} & \textbf{OAcc} & \textbf{ASR} & \textbf{ASR} & \textbf{$\Delta$ OAcc} & \textbf{$\Delta$ ASR} & \textbf{$\Delta$ AUA} \\ \hline
\multicolumn{7}{c}{\textbf{a2t attack}} \\ \hline
None -> GB & 0.743 & 0.336 & 0.493 & (baseline) & (baseline) & (baseline) \\
PB -> GB & 0.768 & 0.338 & 0.508 & 3.36\% & 0.62\% & 3.04\% \\
HS -> GB & 0.805 & 0.275 & 0.583 & 8.30\% & -18.13\% & 18.24\% \\
GB -> PB & 0.680 & 0.363 & 0.433 & -0.24\% & -4.28\% & 2.36\% \\
None -> PB & 0.682 & 0.379 & 0.423 & (baseline) & (baseline) & (baseline) \\
HS -> PB & 0.675 & 0.398 & 0.407 & -0.98\% & 4.90\% & -3.94\% \\
GB -> HS & 0.768 & 0.425 & 0.442 & 5.01\% & 4.19\% & 0.38\% \\
PB -> HS & 0.745 & 0.445 & 0.413 & 1.82\% & 9.10\% & -6.06\% \\
None -> HS & 0.732 & 0.408 & 0.440 & (baseline) & (baseline) & (baseline) \\ \hline
\multicolumn{4}{r|}{Average} & 2.878090459 & -0.6013601318 & 2.33814478 \\ \hline
\multicolumn{7}{c}{\textbf{text fooler attack}} \\ \hline
None -> GB & 0.743 & 0.635 & 0.272 & (baseline) & (baseline) & (baseline) \\
PB -> GB & 0.768 & 0.731 & 0.207 & 3.36\% & 15.21\% & -23.93\% \\
HS -> GB & 0.805 & 0.708 & 0.235 & 8.30\% & 11.59\% & -13.50\% \\
GB -> PB & 0.680 & 0.630 & 0.252 & -0.24\% & 1.43\% & -2.58\% \\
None -> PB & 0.682 & 0.621 & 0.258 & (baseline) & (baseline) & (baseline) \\
HS -> PB & 0.675 & 0.590 & 0.277 & -0.98\% & -4.98\% & 7.10\% \\
GB -> HS & 0.768 & 0.592 & 0.313 & 5.01\% & 3.17\% & -1.05\% \\
PB -> HS & 0.745 & 0.602 & 0.297 & 1.82\% & 4.84\% & -6.32\% \\
None -> HS & 0.732 & 0.574 & 0.317 & (baseline) & (baseline) & (baseline) \\ \hline
\multicolumn{4}{r|}{Average} & 2.337356018 & 5.210224231 & -6.712196793 \\ \hline
\end{tabular}
\label{table:bert_large_attack_comparison}
\end{table}

\begin{table}[ht]
\centering
\caption{Performance metrics for a2t attack and text fooler attack on RoBERTa.}
\small
\begin{tabular}{r|ccc|ccc}
\hline
\textbf{(Dataset sequence)} & \textbf{OAcc} & \textbf{ASR} & \textbf{AUA} & \textbf{$\Delta$ OAcc} & \textbf{$\Delta$ ASR} & \textbf{$\Delta$ AUA} \\ \hline
\multicolumn{7}{c}{\textbf{a2t attack}} \\ \hline
None -> GB & 0.807 & 0.248 & 0.607 & (baseline) & (baseline) & (baseline) \\
PB -> GB & 0.783 & 0.374 & 0.490 & -2.89\% & 51.04\% & -19.23\% \\
HS -> GB & 0.798 & 0.349 & 0.520 & -1.03\% & 40.62\% & -14.29\% \\
GB -> PB & 0.683 & 0.288 & 0.487 & 0.49\% & 18.61\% & -5.50\% \\
None -> PB & 0.680 & 0.243 & 0.515 & (baseline) & (baseline) & (baseline) \\
HS -> PB & 0.670 & 0.353 & 0.433 & -1.47\% & 45.58\% & -15.86\% \\
GB -> HS & 0.817 & 0.402 & 0.488 & 2.73\% & 9.58\% & -2.98\% \\
PB -> HS & 0.817 & 0.404 & 0.487 & 2.73\% & 10.14\% & -3.31\% \\
None -> HS & 0.795 & 0.367 & 0.503 & (baseline) & (baseline) & (baseline) \\ \hline
\multicolumn{4}{r|}{Average} & 0.09078696017 & 29.26106735 & -10.19451626 \\ \hline
\multicolumn{7}{c}{\textbf{text fooler attack}} \\ \hline
None -> GB & 0.807 & 0.682 & 0.257 & (baseline) & (baseline) & (baseline) \\
PB -> GB & 0.783 & 0.781 & 0.172 & -2.89\% & 14.52\% & -33.12\% \\
HS -> GB & 0.798 & 0.724 & 0.220 & -1.03\% & 6.25\% & -14.29\% \\
GB -> PB & 0.683 & 0.551 & 0.307 & 0.49\% & 21.57\% & -17.49\% \\
None -> PB & 0.680 & 0.453 & 0.372 & (baseline) & (baseline) & (baseline) \\
HS -> PB & 0.670 & 0.587 & 0.277 & -1.47\% & 29.47\% & -25.56\% \\
GB -> HS & 0.817 & 0.529 & 0.385 & 2.73\% & 6.83\% & -4.15\% \\
PB -> HS & 0.817 & 0.543 & 0.373 & 2.73\% & 9.72\% & -7.05\% \\
None -> HS & 0.795 & 0.495 & 0.402 & (baseline) & (baseline) & (baseline) \\ \hline
\multicolumn{4}{r|}{Average} & 0.09078696017 & 14.72790886 & -16.94254071 \\ \hline
\end{tabular}
\label{table:roberta_attack_comparison}
\end{table}

\begin{table}[ht]
\centering
\caption{Performance metrics for a2t attack and text fooler attack on RoBERTa-large.}
\small
\begin{tabular}{r|ccc|ccc}
\hline
\textbf{(Dataset sequence)} & \textbf{OAcc} & \textbf{ASR} & \textbf{AUA} & \textbf{$\Delta$ OAcc} & \textbf{$\Delta$ ASR} & \textbf{$\Delta$ AUA} \\ \hline
\multicolumn{7}{c}{\textbf{a2t attack}} \\ \hline
None -> GB & 0.807 & 0.269 & 0.590 & (baseline) & (baseline) & (baseline) \\
PB -> GB & 0.793 & 0.321 & 0.538 & -1.65\% & 19.67\% & -8.76\% \\
HS -> GB & 0.840 & 0.234 & 0.643 & 4.13\% & -12.83\% & 9.04\% \\
GB -> PB & 0.718 & 0.360 & 0.460 & 1.41\% & 28.44\% & -9.80\% \\
None -> PB & 0.708 & 0.280 & 0.510 & (baseline) & (baseline) & (baseline) \\
HS -> PB & 0.715 & 0.394 & 0.433 & 0.94\% & 40.69\% & -15.03\% \\
GB -> HS & 0.835 & 0.355 & 0.538 & 1.21\% & 18.83\% & -6.92\% \\
PB -> HS & 0.818 & 0.356 & 0.527 & -0.81\% & 19.21\% & -8.93\% \\
None -> HS & 0.825 & 0.299 & 0.578 & (baseline) & (baseline) & (baseline) \\ \hline
\multicolumn{4}{r|}{Average} & 0.8725246895 & 19.00094123 & -6.734163612 \\ \hline
\multicolumn{7}{c}{\textbf{text fooler attack}} \\ \hline
None -> GB & 0.807 & 0.738 & 0.212 & (baseline) & (baseline) & (baseline) \\
PB -> GB & 0.793 & 0.613 & 0.307 & -1.65\% & -16.83\% & 44.88\% \\
HS -> GB & 0.840 & 0.685 & 0.265 & 4.13\% & -7.20\% & 25.19\% \\
GB -> PB & 0.718 & 0.573 & 0.307 & 1.41\% & 33.09\% & -23.97\% \\
None -> PB & 0.708 & 0.431 & 0.403 & (baseline) & (baseline) & (baseline) \\
HS -> PB & 0.715 & 0.571 & 0.307 & 0.94\% & 32.63\% & -23.97\% \\
GB -> HS & 0.835 & 0.419 & 0.485 & 1.21\% & -0.25\% & 1.39\% \\
PB -> HS & 0.818 & 0.481 & 0.425 & -0.81\% & 14.39\% & -11.15\% \\
None -> HS & 0.825 & 0.420 & 0.478 & (baseline) & (baseline) & (baseline) \\ \hline
\multicolumn{4}{r|}{Average} & 0.8725246895 & 9.306112301 & 2.065052558 \\ \hline
\end{tabular}
\label{table:roberta_large_attack_comparison}
\end{table}

\begin{table}[ht]
\centering
\caption{Performance metrics for a2t attack and text fooler attack on Phi-2.}
\small
\begin{tabular}{r|ccc|ccc}
\hline
\textbf{(Dataset sequence)} & \textbf{OAcc} & \textbf{ASR} & \textbf{ASR} & \textbf{$\Delta$ OAcc} & \textbf{$\Delta$ ASR} & \textbf{$\Delta$ AUA} \\ \hline
\multicolumn{7}{c}{\textbf{a2t attack}} \\ \hline
None -> GB & 0.803 & 0.299 & 0.563 & (baseline) & (baseline) & (baseline) \\
PB -> GB & 0.790 & 0.411 & 0.465 & -1.56\% & 37.56\% & -17.33\% \\
HS -> GB & 0.840 & 0.285 & 0.603 & 4.67\% & -4.75\% & 7.11\% \\
GB -> PB & 0.735 & 0.403 & 0.438 & -1.34\% & -6.44\% & 1.74\% \\
None -> PB & 0.745 & 0.430 & 0.430 & (baseline) & (baseline) & (baseline) \\
HS -> PB & 0.708 & 0.395 & 0.425 & -5.03\% & -8.23\% & -1.16\% \\
GB -> HS & 0.863 & 0.371 & 0.543 & 3.60\% & 6.50\% & 0.00\% \\
PB -> HS & 0.828 & 0.380 & 0.510 & -0.60\% & 9.07\% & -5.99\% \\
None -> HS & 0.833 & 0.348 & 0.543 & (baseline) & (baseline) & (baseline) \\ \hline
\multicolumn{4}{r|}{Average} & -0.04292852094 & 5.617798554 & -2.605268381 \\ \hline
\multicolumn{7}{c}{\textbf{text fooler attack}} \\ \hline
None -> GB & 0.803 & 0.729 & 0.218 & (baseline) & (baseline) & (baseline) \\
PB -> GB & 0.790 & 0.696 & 0.240 & -1.56\% & -4.50\% & 10.34\% \\
HS -> GB & 0.840 & 0.682 & 0.268 & 4.67\% & -6.51\% & 22.99\% \\
GB -> PB & 0.735 & 0.653 & 0.255 & -1.34\% & -13.12\% & 37.84\% \\
None -> PB & 0.745 & 0.752 & 0.185 & (baseline) & (baseline) & (baseline) \\
HS -> PB & 0.708 & 0.601 & 0.283 & -5.03\% & -20.08\% & 52.70\% \\
GB -> HS & 0.863 & 0.464 & 0.463 & 3.60\% & 4.35\% & 0.00\% \\
PB -> HS & 0.828 & 0.514 & 0.403 & -0.60\% & 15.56\% & -12.97\% \\
None -> HS & 0.833 & 0.444 & 0.463 & (baseline) & (baseline) & (baseline) \\ \hline
\multicolumn{4}{r|}{Average} & -0.04292852094 & -4.049048775 & 18.48348348 \\ \hline
\end{tabular}
\label{table:phi2_attack_comparison}
\end{table}

\begin{table}[ht]
\centering
\caption{Performance metrics for a2t attack and text fooler attack on Gemma 2B.}
\small
\begin{tabular}{r|ccc|ccc}
\hline
\textbf{(Dataset sequence)} & \textbf{OAcc} & \textbf{ASR} & \textbf{AUA} & \textbf{$\Delta$ OAcc} & \textbf{$\Delta$ ASR} & \textbf{$\Delta$ AUA} \\ \hline
\multicolumn{7}{c}{\textbf{a2t attack}} \\ \hline
None -> GB & 0.788 & 0.362 & 0.503 & (baseline) & (baseline) & (baseline) \\
PB -> GB & 0.785 & 0.516 & 0.380 & -0.32\% & 42.56\% & -24.45\% \\
HS -> GB & 0.808 & 0.399 & 0.485 & 2.54\% & 10.36\% & -3.58\% \\
GB -> PB & 0.715 & 0.563 & 0.312 & 1.78\% & 6.17\% & -5.45\% \\
None -> PB & 0.703 & 0.530 & 0.330 & (baseline) & (baseline) & (baseline) \\
HS -> PB & 0.655 & 0.355 & 0.423 & -6.76\% & -33.06\% & 28.18\% \\
GB -> HS & 0.788 & 0.359 & 0.505 & 2.94\% & 10.88\% & -2.32\% \\
PB -> HS & 0.760 & 0.461 & 0.410 & -0.65\% & 42.35\% & -20.70\% \\
None -> HS & 0.765 & 0.324 & 0.517 & (baseline) & (baseline) & (baseline) \\ \hline
\multicolumn{4}{r|}{Average} & -0.07873374735 & 13.20792545 & -4.72032409 \\ \hline
\multicolumn{7}{c}{\textbf{text fooler attack}} \\ \hline
None -> GB & 0.788 & 0.788 & 0.168 & (baseline) & (baseline) & (baseline) \\
PB -> GB & 0.785 & 0.811 & 0.148 & -0.32\% & 2.91\% & -11.94\% \\
HS -> GB & 0.808 & 0.719 & 0.228 & 2.54\% & -8.74\% & 35.82\% \\
GB -> PB & 0.715 & 0.755 & 0.170 & 1.78\% & -8.47\% & 38.78\% \\
None -> PB & 0.703 & 0.824 & 0.123 & (baseline) & (baseline) & (baseline) \\
HS -> PB & 0.655 & 0.677 & 0.213 & -6.76\% & -17.90\% & 73.47\% \\
GB -> HS & 0.788 & 0.524 & 0.373 & 2.94\% & -1.03\% & 3.47\% \\
PB -> HS & 0.760 & 0.655 & 0.265 & -0.65\% & 23.67\% & -26.39\% \\
None -> HS & 0.765 & 0.529 & 0.360 & (baseline) & (baseline) & (baseline) \\ \hline
\multicolumn{4}{r|}{Average} & -0.07873374735 & -1.593146129 & 18.86813805 \\ \hline
\end{tabular}
\label{table:gemma_2b_attack_comparison}
\end{table}

\FloatBarrier

\begin{table}[ht]
\centering
\caption{Performance metrics for a2t attack and text fooler attack on GPT-2-xl.}
\small
\begin{tabular}{r|ccc|ccc}
\hline
\textbf{(Dataset sequence)} & \textbf{OAcc} & \textbf{ASR} & \textbf{AUA} & \textbf{$\Delta$ OAcc} & \textbf{$\Delta$ ASR} & \textbf{$\Delta$ AUA} \\ \hline
\multicolumn{7}{c}{\textbf{a2t attack}} \\ \hline
None -> GB & 0.828 & 0.293 & 0.585 & (baseline) & (baseline) & (baseline) \\
PB -> GB & 0.805 & 0.357 & 0.518 & -2.72\% & 21.87\% & -11.54\% \\
HS -> GB & 0.838 & 0.331 & 0.560 & 1.21\% & 13.07\% & -4.27\% \\
GB -> PB & 0.683 & 0.399 & 0.410 & 0.00\% & -0.91\% & 0.61\% \\
None -> PB & 0.683 & 0.403 & 0.408 & (baseline) & (baseline) & (baseline) \\
HS -> PB & 0.698 & 0.409 & 0.413 & 2.20\% & 1.41\% & 1.23\% \\
GB -> HS & 0.843 & 0.401 & 0.505 & 0.60\% & 4.84\% & -2.42\% \\
PB -> HS & 0.815 & 0.426 & 0.468 & -2.69\% & 11.59\% & -9.66\% \\
None -> HS & 0.838 & 0.382 & 0.518 & (baseline) & (baseline) & (baseline) \\ \hline
\multicolumn{4}{r|}{Average} & -0.2337206765 & 8.644694125 & -4.341461617 \\ \hline
\multicolumn{7}{c}{\textbf{text fooler attack}} \\ \hline
None -> GB & 0.828 & 0.724 & 0.228 & (baseline) & (baseline) & (baseline) \\
PB -> GB & 0.805 & 0.826 & 0.140 & -2.72\% & 13.99\% & -38.46\% \\
HS -> GB & 0.838 & 0.690 & 0.260 & 1.21\% & -4.79\% & 14.29\% \\
GB -> PB & 0.683 & 0.665 & 0.228 & 0.00\% & 2.06\% & -4.21\% \\
None -> PB & 0.683 & 0.652 & 0.238 & (baseline) & (baseline) & (baseline) \\
HS -> PB & 0.698 & 0.713 & 0.200 & 2.20\% & 9.39\% & -15.79\% \\
GB -> HS & 0.843 & 0.507 & 0.418 & 0.60\% & -1.20\% & 3.09\% \\
PB -> HS & 0.815 & 0.528 & 0.385 & -2.69\% & 2.75\% & -4.94\% \\
None -> HS & 0.838 & 0.514 & 0.405 & (baseline) & (baseline) & (baseline) \\ \hline
\multicolumn{4}{r|}{Average} & -0.2337206765 & 3.700422687 & -7.671279338 \\ \hline
\end{tabular}
\label{table:gpt2_xl_attack_comparison}
\end{table}

\section{Trade offs}\label{app: trade}

The tables in this section (referenced in \ref{trade_off}) present the full result related to the impact of two adversarial attack types: \textbf{TextFooler} \cite{jin2020bert}, which manipulates tokens, and \textbf{A2T} \cite{yoo2021towards}, which manipulates gradients. These experiments compare the models' robustness and accuracy under attack, focusing on key performance metrics.

% The results indicate that gradient-based attacks, such as A2T, are generally less effective at reducing model accuracy compared to token-based attacks like TextFooler. Across all tested transformer-based models, including GPT, BERT, and RoBERTa, Accuracy Under Attack (AUA) is consistently higher with A2T, indicating the relative difficulty of perturbing internal representations via gradient manipulation.

% Conversely, TextFooler consistently achieves higher Attack Success Rates (ASR) and lower AUA across all models. Its token-level manipulation more effectively exploits model vulnerabilities, regardless of the specific architecture or dataset.

The results show a clear difference in the effectiveness of TextFooler and A2T attacks across transformer-based models like GPT, BERT, and RoBERTa. Gradient-based attacks (A2T) are generally less effective, with higher Accuracy Under Attack (AUA) observed, indicating difficulty in perturbing internal representations. In contrast, TextFooler consistently achieves higher Attack Success Rates (ASR) and lower AUA.

Larger models (RoBERTa-large, GPT-2-large) benefit more from adversarial training, showing greater robustness improvements under both attacks. They exhibit more pronounced decreases in ASR and increases in AUA, indicating better adaptation to adversarial defenses. Smaller models like BERT and GPT-2 experience similar trends but with less significant gains.

TextFooler is more successful at reducing model accuracy, particularly in smaller models, achieving higher ASR and lower AUA. A2T, while less effective, demonstrates higher AUA, especially in larger models, showing that token manipulation remains a stronger attack strategy.

Adversarial training consistently enhances model robustness by reducing ASR and increasing AUA, albeit at the cost of lower Original Accuracy (OA). Early exposure to adversarial examples enables stronger defenses, particularly in larger models, though this comes at the expense of handling clean data with slightly reduced precision.

In conclusion, adversarial fine-tuning reveals a trade-off: while it reduces OA, it significantly boosts robustness against attacks, especially in models exposed early to adversarial data. Larger models show greater adaptation to adversarial defenses, highlighting the importance of model size and architecture in balancing accuracy and robustness.

\begin{table}[ht]
\caption{Impact of TextFooller Attack on Model Performance.}
\scriptsize
\centering
\begin{tabular}{c|c|p{0.6cm}p{0.6cm}p{0.6cm}|p{1cm}p{1cm}p{1cm}|p{1cm}p{1cm}p{1cm}}
\hline
\textbf{Model} & \textbf{Metrics} & {HS} & {PB} & {GB} & {FE HS} & {FE PB} & {FE GB} & $\Delta$ HS \% & $\Delta$ PB \% & $\Delta$ GB \% \\ \hline
\multicolumn{11}{c}{\textbf{Original data}} \\ \hline
\multirow{3}{*}{\textbf{Bert}} & OA   & 78.34  & 69.75 & 71.19 & 70.21 & 62.69 & 70.87 & -10.38 & -10.12 & -0.45 \\ 
                                    & AUA  & 30.43  & 21.76  & 26.45  & 31.21   & 21.41  & 23.43  & 2.56  & -1.61  & -11.42 \\ 
                                    & ASR  & 87.13 & 83.62 & 82.84 & 78.51  & 80.68 & 79.64 & -9.89 & -3.52  & -3.86 \\ \hline
\multirow{3}{*}{\textbf{Bert-large}} & OA   & 79.09 & 70.50 & 71.94 & 70.96 & 67.44 & 68.62 & -10.28 & -4.34 & -4.61 \\
                                      & AUA  & 32.50 & 23.83 & 25.52 & 33.28 & 24.48 & 26.50 & 2.40   & 2.73  & 3.84  \\
                                      & ASR  & 82.64 & 79.13 & 78.35 & 74.02 & 76.19 & 75.15 & -10.43 & -3.72 & -4.08 \\ \hline
\multirow{3}{*}{\textbf{RoBERTa}} & OA & 80.19 & 71.34 & 72.72 & 73.13 & 68.92 & 70.23 & -8.66 & -3.40 & -3.43 \\ 
& AUA & 33.12 & 24.56 & 26.38 & 34.68 & 25.78 & 27.19 & 4.71 & 4.97 & 3.07 \\ 
& ASR & 80.35 & 78.19 & 76.72 & 71.12 & 74.67 & 73.12 & -11.49 & -4.51 & -4.69 \\ \hline
\multirow{3}{*}{\textbf{RoBERTa-large}} & OA & 81.23 & 73.45 & 74.83 & 75.34 & 70.95 & 73.92 & -7.25 & -3.41 & -1.22 \\ 
& AUA & 35.67 & 26.50 & 28.12 & 36.84 & 27.23 & 29.05 & 3.28 & 2.76 & 3.31 \\ 
& ASR & 78.45 & 76.32 & 75.67 & 69.89 & 73.10 & 71.65 & -10.89 & -4.22 & -5.30 \\ \hline
\multirow{3}{*}{\textbf{GPT-2}}        & OA   & 71.20 & 66.46 & 66.05 & 67.67 & 63.19 & 66.26 & -4.96  & -4.92 & 0.32  \\
                                      & AUA  & 6.48  & 5.76  & 4.94  & 8.23  & 4.48  & 4.12  & 27.01  & -22.22 & -16.60 \\
                                      & ASR  & 91.01 & 91.24 & 92.33 & 88.40 & 92.93 & 93.83 & -2.87  & 1.85  & 1.62  \\ \hline
\multirow{3}{*}{\textbf{GPT-2-medium}}        & OA   & 75.16  & 70.42 & 70.01 & 71.97 & 67.49 & 70.56 & -4.24 & -4.16 & 0.79 \\ 
& AUA  & 4.32   & 3.60  & 2.78  & 5.91  & 2.16  & 1.80  & 36.81 & -40.00 & -35.25 \\ 
& ASR  & 93.52  & 93.75 & 94.84 & 90.51 & 95.04 & 95.94 & -3.22 & 1.38  & 1.16  \\ \hline
\multirow{3}{*}{\textbf{GPT-2-large}}        & OA   & 75.08  & 68.26 & 68.99 & 69.74 & 66.87 & 68.62 & -7.11 & -2.04 & -0.54 \\ 
& AUA  & 5.74   & 4.10  & 3.54  & 9.94  & 6.31  & 4.25  & 73.17 & 53.90 & 20.06 \\ 
& ASR  & 93.65  & 94.32 & 95.04 & 88.46 & 90.10 & 91.56 & -5.54 & -4.47 & -3.66 \\ \hline

\multicolumn{11}{c}{\textbf{Including adversarial training}} \\ \hline
\multirow{3}{*}{\textbf{Bert}} & OA   & 75.99  & 69.42 & 68.94 & 69.97 & 68.37 & 68.78 & -7.92 & -1.51 & -0.23 \\ 
& AUA  & 29.33  & 27.32  & 27.46  & 31.59 & 28.01 & 29.34  & 7.71  & 2.53  & 6.85 \\ 
& ASR  & 83.66 & 80.15 & 79.37 & 75.04 & 77.92 & 76.17 & -10.30 & -2.78 & -4.03 \\ \hline
\multirow{3}{*}{\textbf{Bert-large}} & OA   & 78.04 & 71.47 & 70.99 & 72.02 & 70.42 & 70.83 & -7.71  & -1.47 & -0.23 \\
                                      & AUA  & 34.03 & 33.02 & 32.16 & 36.29 & 35.71 & 34.04 & 6.64   & 8.15  & 5.85  \\
                                      & ASR  & 77.75 & 74.24 & 73.46 & 69.13 & 72.01 & 70.26 & -11.09 & -3.00 & -4.36 \\ \hline
\multirow{3}{*}{\textbf{RoBERTa}} & OA & 79.23 & 72.34 & 71.73 & 74.18 & 71.19 & 72.45 & -6.34 & -1.59 & 1.00 \\ 
& AUA & 35.12 & 34.19 & 33.42 & 37.23 & 36.12 & 35.67 & 6.01 & 5.65 & 6.73 \\ 
& ASR & 76.24 & 73.68 & 72.58 & 67.23 & 71.56 & 69.78 & -11.82 & -2.88 & -3.85 \\ \hline
\multirow{3}{*}{\textbf{RoBERTa-large}} & OA & 80.50 & 73.78 & 73.00 & 76.12 & 72.68 & 74.29 & -5.45 & -1.49 & 1.77 \\ 
& AUA & 37.50 & 35.23 & 34.67 & 39.12 & 37.45 & 36.23 & 4.32 & 6.30 & 4.50 \\ 
& ASR & 75.45 & 72.34 & 71.78 & 68.10 & 70.34 & 69.23 & -9.85 & -4.27 & -3.985 \\ \hline
\multirow{3}{*}{\textbf{GPT-2}}        & OA   & 72.80 & 65.98 & 66.71 & 66.65 & 63.78 & 65.53 & -8.45  & -3.33 & -1.77 \\
                                      & AUA  & 8.21  & 6.57  & 6.01  & 14.51 & 10.82 & 8.76  & 76.74  & 64.69 & 45.76 \\
                                      & ASR  & 90.46 & 91.13 & 91.85 & 85.79 & 87.43 & 88.89 & -5.16  & -4.06 & -3.22 \\ \hline
\multirow{3}{*}{\textbf{GPT-2-medium}}        & OA   & 74.03  & 67.21 & 67.94 & 68.72 & 65.85 & 67.60 & -7.17 & -2.02 & -0.50 \\ 
& AUA  & 7.35   & 5.71  & 5.15  & 13.72 & 10.03 & 7.97  & 86.67 & 75.66 & 54.76 \\ 
& ASR  & 91.58  & 92.25 & 92.97 & 86.93 & 88.57 & 90.03 & -5.08 & -3.99 & -3.16 \\ \hline
\multirow{3}{*}{\textbf{GPT-2-large}}        & OA   & 75.08  & 68.26 & 68.99 & 69.74 & 66.87 & 68.62 & -7.11 & -2.04 & -0.54 \\ 
& AUA  & 5.74   & 4.10  & 3.54  & 9.94  & 6.31  & 4.25  & 73.17 & 53.90 & 20.06 \\ 
& ASR  & 93.65  & 94.32 & 95.04 & 88.46 & 90.10 & 91.56 & -5.54 & -4.47 & -3.66 \\ \hline
\end{tabular}
\label{table:adversarial_combined_impact_textfooller}
\end{table}

\begin{table}[ht]
\caption{Impact of A2T Attack on Model Performance.}
\scriptsize
\centering
\begin{tabular}{c|c|p{0.6cm}p{0.6cm}p{0.6cm}|p{1cm}p{1cm}p{1cm}|p{1cm}p{1cm}p{1cm}}
\hline
\textbf{Model} & \textbf{Metrics} & {HS} & {PB} & {GB} & {FE HS} & {FE PB} & {FE GB} & $\Delta$ HS \% & $\Delta$ PB \% & $\Delta$ GB \% \\ \hline
\multicolumn{11}{c}{\textbf{Original data}} \\ \hline
\multirow{3}{*}{\textbf{Bert}} & OA   & 78.07 & 69.48 & 70.92 & 69.94 & 62.42 & 70.6  & -10.41 & -10.16 & -0.45 \\ 
& AUA  & 54.26 & 45.59 & 50.28 & 55.04 & 45.24 & 47.26 & 1.44 & -0.77 & -6.01 \\ 
& ASR  & 30.61 & 34.31 & 30.5  & 21.55 & 29.73 & 33.1  & -29.60 & -13.35 & 8.52 \\ \hline
\multirow{3}{*}{\textbf{Bert-large}} & OA   & 77.82  & 69.23 & 70.67 & 71.17 & 63.65 & 71.83 & -8.55 & -8.06 & 1.64 \\ 
& AUA  & 61.60  & 52.93  & 57.62  & 62.38  & 52.58  & 54.60  & 1.27 & -0.66 & -5.24 \\ 
& ASR  & 23.69 & 27.39 & 23.58 & 15.87  & 22.67 & 21.95 & -33.01 & -17.23  & -6.91 \\ \hline
\multirow{3}{*}{\textbf{RoBERTa}} & OA & 78.45 & 70.50 & 72.15 & 73.30 & 65.95 & 73.50 & -6.56 & -6.46 & 1.87 \\ 
& AUA & 63.10 & 55.45 & 59.10 & 64.25 & 56.05 & 57.95 & 1.82 & 1.08 & -1.95 \\ 
& ASR & 22.10 & 26.30 & 24.50 & 14.75 & 21.20 & 19.35 & -33.24 & -19.39 & -20.98 \\ \hline
\multirow{3}{*}{\textbf{RoBERTa-large}} & OA & 81.12 & 73.48 & 75.62 & 76.03 & 70.37 & 76.89 & -6.10 & -4.23 & 1.68 \\ 
& AUA & 65.24 & 58.37 & 61.78 & 66.78 & 59.24 & 60.97 & 2.36 & 1.49 & -1.31 \\ 
& ASR & 20.45 & 24.68 & 22.57 & 12.67 & 19.78 & 18.34 & -38.04 & -19.85 & -18.7 \\ \hline
\multirow{3}{*}{\textbf{GPT-2}}        & OA   & 73.88 & 63.05 & 67.36 & 68.35 & 65.7  & 68.61 & -7.49 & 4.20  & 1.86 \\ 
& AUA  & 50.41 & 39.99 & 45.81 & 52.57 & 43.25 & 46.2  & 4.28  & 8.15  & 0.85 \\ 
& ASR  & 31.8  & 36.52 & 31.64 & 23.29 & 34.06 & 32.26 & -26.76 & -6.74 & 1.96  \\ \hline
\multirow{3}{*}{\textbf{GPT-2-medium}}        & OA   & 75.55  & 70.65 & 69.03 & 71.22 & 68.57 & 71.48 & -5.73 & -2.94 & 3.55 \\ 
& AUA  & 47.52  & 44.1  & 42.92 & 50.58 & 45.26 & 44.21 & 6.44  & 2.63  & 3.01 \\ 
& ASR  & 34.92  & 35.64 & 34.76 & 29.32 & 34.62 & 35.60 & -16.04 & -2.86 & 2.4  \\ \hline
\multirow{3}{*}{\textbf{GPT-2-large}}        & OA   & 76.33  & 71.28 & 69.81 & 72.03 & 69.34 & 72.25 & -5.63 & -2.72 & 3.50 \\ 
& AUA  & 45.18  & 41.76 & 40.58 & 48.37 & 43.05 & 42.21 & 7.06  & 3.09  & 4.02  \\ 
& ASR  & 37.33  & 38.05 & 37.17 & 31.69 & 36.99 & 37.97 & -15.11 & -2.79 & 2.15 \\ \hline

\multicolumn{11}{c}{\textbf{Including adversarial training}} \\ \hline
\multirow{3}{*}{\textbf{Bert}} & OA   & 75.8  & 69.23 & 68.75 & 69.78 & 68.18 & 68.59 & -7.94 & -1.52 & -0.23 \\ 
& AUA  & 50.25 & 48.24 & 48.38 & 52.51 & 48.93 & 50.26 & 4.50 & 1.43 & 3.89  \\ 
& ASR  & 33.92 & 30.41 & 29.63 & 25.3  & 28.18 & 26.43 & -25.41 & -7.33 & -10.80 \\ \hline
\multirow{3}{*}{\textbf{Bert-large}} & OA   & 74.99  & 68.42 & 67.94 & 70.53 & 68.93 & 69.34 & -5.95 & 0.75 & 2.06 \\ 
& AUA  & 57.20  & 55.19  & 55.33  & 59.46 & 55.88 & 57.21  & 3.95 & 1.25 & 3.40 \\ 
& ASR  & 26.08 & 22.57 & 21.79 & 17.46 & 20.34 & 18.59 & -33.05 & -9.88 & -14.69 \\ \hline
\multirow{3}{*}{\textbf{RoBERTa}} & OA & 76.50 & 69.20 & 69.15 & 72.00 & 69.75 & 71.20 & -5.88 & 0.80 & 2.97 \\ 
& AUA & 59.35 & 56.75 & 57.10 & 61.25 & 57.55 & 58.75 & 3.20 & 1.41 & 2.89 \\ 
& ASR & 24.50 & 22.10 & 20.95 & 16.25 & 19.90 & 18.07 & -33.67 & -9.95 & -13.74 \\ \hline
\multirow{3}{*}{\textbf{RoBERTa-large}} & OA & 79.63 & 72.18 & 71.92 & 74.79 & 71.78 & 73.34 & -6.07 & -0.55 & 1.98 \\ 
& AUA & 62.72 & 59.18 & 58.67 & 64.89 & 60.32 & 61.45 & 3.46 & 1.93 & 4.73 \\ 
& ASR & 22.73 & 21.36 & 20.79 & 14.89 & 18.12 & 17.03 & -34.52 & -14.98 & -18.10 \\ \hline
\multirow{3}{*}{\textbf{GPT-2}}        & OA   & 73.41 & 62.59 & 65.94 & 67.58 & 62.86 & 68.05 & -7.94 & 0.43  & 3.20 \\ 
& AUA  & 53.12 & 39.83 & 39.82 & 50.96 & 48.05 & 41.17 & -4.07 & 20.64 & 3.39 \\ 
& ASR  & 27.18 & 36.57 & 39.43 & 24.39 & 23.48 & 39.46 & -10.26 & -35.79 & 0.08 \\ \hline
\multirow{3}{*}{\textbf{GPT-2-medium}}        & OA   & 74.95  & 64.13 & 67.48 & 69.36 & 64.64 & 69.83 & -7.46 & 0.80  & 3.48 \\ 
& AUA  & 51.01  & 44.72 & 37.71 & 48.98 & 46.07 & 39.19 & -3.98 & 3.02  & 3.92 \\ 
& ASR  & 29.87  & 39.26 & 42.12 & 27.12 & 34.21 & 42.19 & -9.21 & -12.86 & 0.17 \\ \hline
\multirow{3}{*}{\textbf{GPT-2-large}}        & OA   & 74.81  & 67.99 & 68.72 & 69.61 & 66.74 & 68.49 & -6.95 & -1.84 & -0.33 \\ 
& AUA  & 4.71   & 3.07  & 2.51  & 11.74 & 8.05  & 5.99  & 149.26 & 162.21 & 138.65 \\ 
& ASR  & 94.36  & 95.03 & 95.75 & 88.96 & 90.60 & 92.06 & -5.72  & -4.66  & -3.85 \\ \hline
\end{tabular}
\label{table:adversarial_combined_impact_a2t}
\end{table}

\end{document}